\documentclass{article}
\usepackage{spconf,amsmath,graphicx}
\usepackage{enumitem}
\setlist{nosep, leftmargin=14pt}
\usepackage{makecell}
\usepackage{mwe} % to get dummy images
\usepackage{amsfonts,amssymb}
\title{Region and Spatial Aware Anomaly Detection for Fundus Images}
\name{Jingqi Niu$^1$,  Shiwen Dong$^1$ ,  Qinji Yu$^1$, Kang Dang$^2$$^*$, Xiaowei Ding$ ^{1,2}$$^*$ \thanks{* Corresponding Authors}}
\address{$^1$Shanghai Jiao Tong University, Shanghai, China \\
$^2$Voxelcloud Inc, Shanghai, China
}
\begin{document}
%\ninept
%
\maketitle
\begin{abstract}
Recently anomaly detection has drawn much attention in diagnosing ocular diseases. Most existing anomaly detection research in fundus images has relatively large anomaly scores in the salient retinal structures, such as blood
vessels, optical cups and discs. In this paper,  we propose a Region and Spatial Aware Anomaly Detection (ReSAD) method for fundus images, which obtains  local region and long-range spatial information to reduce the false positives in the normal structure. ReSAD transfers a pre-trained model to extract the features of normal fundus images and applies the Region-and-Spatial-Aware feature Combination module (ReSC) for  pixel-level features to build a memory bank. In the testing phase, ReSAD 
uses the memory bank to determine out-of-distribution samples as abnormalities. Our method significantly outperforms the existing anomaly detection methods for fundus images on two publicly benchmark datasets.
\end{abstract}
\section{Introduction}
\label{sec:intro}

%Patients with ocular diseases such as diabetic retinopathy (DR) and age-related macular degeneration (AMD)  are often unaware of the asymptomatic progression \cite{robinson2003prevalence}. Deep learning has proven its potential for automatic ocular disease diagnosis for fundus images \cite{li2021applications}. However,  training a high-performance deep model requires massive labeled data, which consumes substantial cost and labor. In view of this problem, anomaly detection has drawn increasing attention from the research community \cite{zhang2022multi,zhou2021proxy}. The anomaly detection method, which does not need abnormal data with annotation, is suitable for detecting and diagnosing ocular diseases.

%Deep learning has proven its potential for automatic medical disease diagnosis \cite{shen2017deep}. However,  training a high-performance deep model requires massive labeled abnormal images, which are hard to collect and annotate for rare diseases. Anomaly detection utilizes only normal data for medical diagnosis, which is suitable for rare disease diagnoses. Medical anomaly detection methods are widely used in various medical images\cite{cai2022dual,zhou2021proxy}. Meanwhile, only a few research focus on anomaly detection for small and scattered retinal lesions. Therefore, this paper aims to construct an anomaly detection framework for fundus images.

Deep learning has revolutionized computer-assisted medical disease diagnosis \cite{shen2017deep}. However, existing supervised deep-learning models usually require abundant training data that is difficult to obtain, especially for uncommon diseases. Therefore, we study anomaly detection problems, relying only on a normal dataset to construct a model representing healthy subjects and identify any abnormalities in the patients. Anomaly detection is a widely studied topic in various medical image analysis applications \cite{cai2022dual,zhou2021proxy}; however, only a few existing works focus on anomaly detection in fundus images. This paper tackles the problem of anomaly detection in color fundus images, which is highly challenging due to the small size and diversity of the lesions.

%Deep learning has proven its potential for automatic medical disease diagnosis for fundus images \cite{li2021applications}. However,  training a high-performance deep model requires massive labeled abnormal images, which consumes substantial cost and labor. Many medical diseases have a long-tail distribution, and rare cases are challenging to collect. Anomaly detection utilizes only normal data for medical diagnosis, which is suitable for diagnosing rare diseases. Medical anomaly detection methods are widely used in X-rays and magnetic resonance imaging (MRI) images, while only a few research on the fundus image for anomaly detection. Aiming at this research vacancy, we propose an anomaly detection framework for fundus images.

%Most existing anomaly detection methods for fundus images use the reconstruction-based approach \cite{zhang2022multi,schlegl2019f}. The method trains a reconstruction model for normal data and detects abnormalities by reconstruction errors. Besides, 

Our work builds on state-of-the-art representative-based anomaly detection approaches. Such approaches map the normal image into a specific feature space in the training phase and detect out-of-distribution samples as abnormalities in model testing. Similar to the previous work \cite{salehi2021multiresolution,roth2022towards}, we directly leverage deep representation from ImageNet classification without additional feature finetuning. To adapt the features for the fundus anomaly detection task, we propose a novel Region-and-Spatial-Aware feature Combination (ReSC) module. Region-aware feature module utilizes distance-related relationships within the local image region, while the spatial-aware feature module constructs long-range spatial relationships across the entire fundus image. Experiments demonstrated that our region-and-spatial-ware feature combination module could significantly reduce false positive detection and improve anomaly detection quality.

The overall procedure of our proposed region-and-spatial-aware
anomaly detection method (ReSAD) is as follows. ReSAD first
adapts a pre-trained ImageNet model to extract the features
from normal fundus images. In order to reduce the false positives,  a region-and-spatial-aware feature combination module is introduced to incorporate local region and long-range spatial
information, and a memory bank is constructed from these features. In the testing phase, ReSAD calculates the distance between the testing features and the features in the memory bank to determine out-of-distribution anomaly samples. 

Contributions: (1) we propose a representation-based region-and-spatial-aware anomaly detection method(ReSAD), which utilizes local region and long-range spatial information to reduce false positives and enhance anomaly detection accuracy. (2) Our method outperformed the state-of-the-art models on the IDRiD and ADAM datasets by 6.2\% and 8.7\%  in pixel-level AUC, respectively. In addition, our method can outperform most state-of-the-art methods only with 10\% of normal images as training data.

\section{related work}
\label{sec:related}
This section discusses the current state-of-the-art anomaly detection works for fundus images.

The reconstruction-based approach is widely used in both natural and medical images for anomaly detection \cite{cai2022dual}. fAno-GAN \cite{schlegl2019f} utilized a generative adversarial model to learn the distribution of normal data by reconstruction and measure the anomaly score by reconstruction errors. WDMT-Net \cite{zhang2022multi} proposed a multi-task encoder-decoder network with weight decay skip connection for anomaly detection. The reconstruction-based approach have relatively large reconstruction errors in normal retinal structures, which causes the false positives. %Besides, the reconstruction-based approach needs enough normal images to reconstruct the normal fundus image, %and insufficient normal data may lead to performance degradation.

Few works explored the representation-based anomaly detection approach for fundus images. Deep-IF \cite{ouardini2019towards} presented a transfer learning method and fitted an Isolation Forest model for anomaly detection. MDFSC \cite{das2022anomaly} applied an autoencoder to create a feature space and adapted the multi-scale sparse coding for anomaly detection. The existing methods have not considered model pixel-level features for fundus images, which led to performance degradation and cannot analyze interpretability for retinal lesions.

Compared with existing methods, ReSAD proposed in this paper is data-efficient and can reduce the false positives in normal structure regions.

\section{Method}
\label{sec:method}
\begin{figure*}[htb]

  \centerline{\includegraphics[width=18cm]{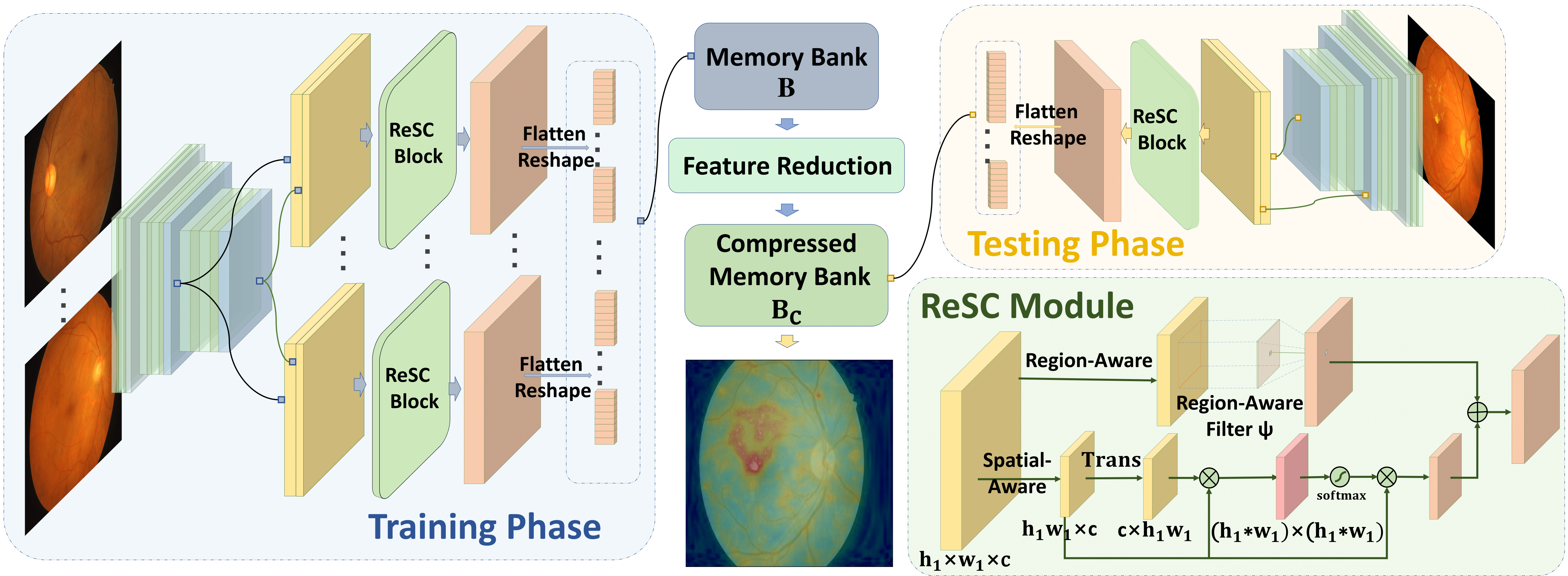}}
\vspace{-0.7em}
\caption{ReSAD training pipeline. Firstly, ReSAD extracts multi-level feature maps through a pre-trained model. Secondly,  the ReSC module introduces  local region and long-range information into the pixel-level features and constructs a memory bank. In the testing phase, ReSAD determines the out-of-distribution samples as abnormalities through the memory bank. }
\vspace{-0.7em}
\label{fig:total}
\end{figure*}

The ReSAD framework consists of three parts:  Feature Extraction  (\S \ref{sec:Feature_Extraction}), Region and Spatial Aware Memory Bank Building (\S \ref{sec:MLFA}) and Inference (\S \ref{sec:inference}). The complete pipeline is shown in Fig.\ref{fig:total}.

\vspace{-1.2em}
\subsection{Feature Extraction}
\label{sec:Feature_Extraction}
In order to detect the abnormalities of the fundus image, it is necessary to construct the distribution of normal data  in the training phase. In this paper, we adapt a model $\phi$ pre-trained on ImageNet to extract features of the normal images in the training set. Following previous work \cite{roth2022towards}, we apply a ResNet-like model as our feature extractor.

We denote the input image as $I_k \in \mathbb{R}^{h \times w \times 3}$ $(k =1,..., N)$, where $N$ denotes the number of training images, $h$ and $w$ denote the height and width of the input image $I_k$. Anomaly detection in fundus images needs fine-grained and high-level features for localizing small and semantic-related retinal lesions, so we extract multi-level feature map $f_{ik} \in \mathbb{R}^{h_i \times w_i \times c_i}$ $(i = 1,2,3,4)$ from the $i_{th}$ stage block $ \phi_i$ of the pre-trained model $ \phi$, which can be calculated by:
\begin{gather}
\label{equ:feature_map}
    f_{ik} = \phi_i(f_{(i-1)k}), \notag \\
    f_{1k} = \phi_1(I_k),
\end{gather}

To control memory consumption and maintain comprehensive feature representation, ReSAD utilizes the intermediate features $f_{2k}$, $f_{3k}$ to construct the multi-level feature map $F_k$. In order to maintain the spatial resolution, ReSAD  upscales feature map $f_{3k}$ to the same spatial shape as $f_{2k}$ and concatenates the feature map $f_{2k}$ and $f_{3k}$ as:
\begin{equation}
\label{equ:concat}
    F_k = \mathrm{concat} (f_{2k}, \mathrm{T}_u(f_{3k})),
\end{equation}
where $F_k \in \mathbb{R}^{h_2 \times w_2 \times c}$, $c =  c_2 + c_3$ and $\mathrm{T}_u$ denotes the upscale transform.

\subsection{Region and Spatial Aware Memory Bank Building}
\label{sec:MLFA}
Certain relationships of fundus images are localized in a particular region, such as the optical cup disc. Some relationships of fundus images are region-independent, and structural correlations are not concentrated in a region, such as the fundus vessels (the correlation between the superior and inferior fundus vessels). 

To construct the relationships in fundus images, ReSAD applies the Region-and-Spatial-Aware feature Combination module (ReSC) to obtain local region and long-range spatial information for the pixel-level features and utilizes those features to construct a memory bank. Region-aware feature module can extract the distance-related relationships in the retinal structure region.   Spatial-aware feature module can construct the spatial relationship of the fundus structures regardless of their distance in the spatial dimension. 

Firstly, ReSC applies a fixed regional distance-aware filter $\psi_r$ to extract local region information,  where $r$ denotes the size of the filter $\psi$. The weights of the filter linearly decay with the distance from the center. We can calculate the region-aware feature map $R_k$ as follows:
\begin{equation}
\label{equ:reg_map}
    R_k = \psi_r(F_k),
\end{equation}
where $R_k \in \mathbb{R}^{h_2 \times w_2 \times c}$ has the same shape as $F_k$.

Secondly, ReSC applies a spatial self-attention operation for the feature map $F_k$ to obtain the spatial-aware feature map $P_k$. ReSC first flattens the spatial dimension of $F_k$ and calculates spatial attention map $G_k = mul(F_k,F_k^\mathsf{T})$, where $mul$ denotes the matrix multiplication and $G_k \in \mathbb{R}^{h_2w_2 \times h_2w_2}$. In spatial attention map $G_k$, every two similar features can contribute improvement regardless the spatial distance between the two features. The spatial-aware features map $P_k$ can be calculated as:
\begin{equation}
\label{equ:pos_map}
    P_k = F_k \circ softmax(G_k),
\end{equation}
where $\circ$ denotes the hadamard product, $P_k \in \mathbb{R}^{h_2 \times w_2 \times c}$ has the same  shape as $F_k$.

Finally, we can calculate $C_k$ by:
\begin{equation}
\label{equ:com_map}
    C_k = P_k + R_k.
\end{equation}

To build a memory bank for every pixel-level feature, ReSAD integrates region-and-spatial aware feature $C^{(x,y)}_k$ together, where $C^{(x,y)}_k$ denotes pixel-level feature in the position $(x,y)$ for normal image $I_k$. By integrating  pixel-level features in all positions and all images in the training set, we can get a region and spatial aware memory bank $B$ as:
% \vspace{-0.em}
\begin{equation}
\label{equ:memory bank}
   B = \bigcup_{k,x,y}  C^{(x,y)}_k, k \in N, x \in h_2, y \in w_2,
\end{equation}
% \vspace{-0.2em}
where $C^{(x,y)}_k$ denotes  pixel-level feature in the position $(x,y)$ for normal image $I_k$. $B$ includes $N \times h_2 \times w_2$ features, and each feature have $c$ channels which contains local region and long-range spatial information.

In addition, to reduce the size of $B$, following \cite{roth2022towards}, ReSAD adapts iterative greedy approximation proposed in \cite{sener2017active} to acquire a compressed memory bank $B_{c}$ that retains most of the information and needs less memory consumption.

\subsection{Inference}

\label{sec:inference}
For a single image $I_{test} \in \mathbb{R}^{h \times w \times 3}$, we can calculate its  region-and-spatial-aware feature map $C_{test}$ by Equ.\ref{equ:com_map}  and calculate the distance for each pixel-level feature of the  testing image $d^{x,y}_{test}$ as:
\vspace{-0.2em}
\begin{equation}
\label{equ:inference}
   d^{x,y}_{test} = min_{b \in B_c} \Vert b -  c_{test}^{x,y} \Vert_2,
\end{equation}
where $b,c_{test}^{x,y} \in \mathbb{R}^{c}$, $b$ denotes the pixel-level feature vector in the compressed memory bank $B_c$, $c_{test}^{x,y}$ denotes  region-and-spatial-aware feature for image $I_{test}$ in the position $(x,y)$. We can get the complete distance map $D_{test}$ by integrating the $d_{test}^{x,y}$ in spatial location.

By interpolating $D_{test}$ to the spatial shape of the original image $I_{test}$, we can get the pixel-level anomaly score map $A_{test} \in \mathbb{R}^{h \times w}$. Besides, We utilize the maximum score of the anomaly score map $A_{test}$ as the image-level anomaly score $a_{test}$.
\section{Experiment}
\label{sec:typestyle}

\subsection{Dataset}
\label{sec:Dataset}
We evaluate our method on two public benchmark datasets, IDRiD and ADAM. 

\textbf{IDRiD}: Indian Diabetic Retinopathy Image Dataset \cite{porwal2020idrid} (IDRiD) contains 134 normal images for training, 32 normal images and 81 abnormal images for testing. Each abnormal image contains the mask of four different lesions, including microaneurysms (MA), soft exudates (SE), hard exudates (EX), and hemorrhages (HE). 

\textbf{ADAM}: The Automatic Detection challenge on
Age-related Macular degeneration dataset (ADAM) \cite{fang2022adam} dataset contains 282 normal images for training and 118 abnormal images for testing. Each abnormal image contains the mask of five different lesions, including drusen, exudate, hemorrhage, scars and other lesions. 

The original fundus images are too huge to construct a memory bank ($4288 \times 2848$ for IDRiD and $ 2124 \times 2056$ for ADAM), so we resize each image into $ 784 \times 784$ for anomaly detection in fundus images.

\begin{figure*}[htb]

  \centerline{\includegraphics[width=16cm]{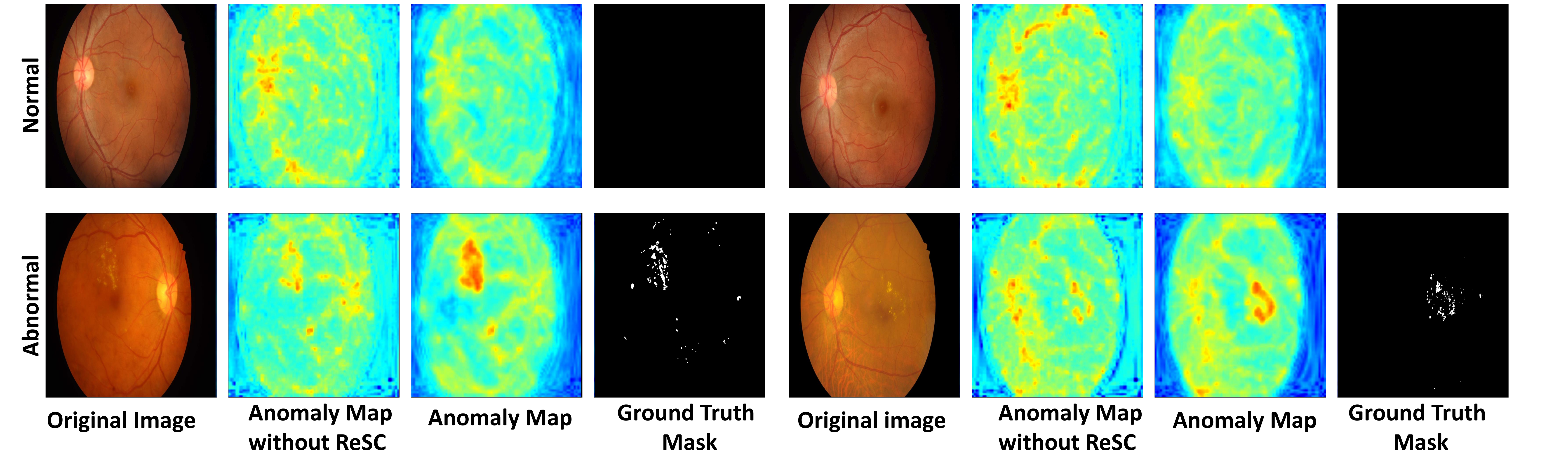}}
 \vspace{-0.3cm}
\caption{The qualitative results of ReSAD on IDRiD dataset.}
 \vspace{-0.3cm}
\label{fig:vis}
\end{figure*}

\vspace{-0.5em}
\subsection{Evaluation Metric}
\label{sec:Metric}
Following previous work \cite{zhou2021proxy,zhang2022multi}, the Area Under Curve (AUC) is adapted to evaluate the performance. Besides, we evaluate the performance of the method via balanced accuracy (ACC) \cite{mower2009prep}, which normalizes true positives and true negatives predictions.
 \vspace{-0.3cm}
\subsection{Implementation Details}
 \vspace{-0.2cm}
\label{sec:Imp_Det}
In our implementation, we adapt WideResnet-50 \cite{zagoruyko2016wide} as the pre-trained ImageNet model. The reduction fraction of the number of training samples is  $0.1$. The weight of the sparse random projection $\epsilon$ is $0.90$. The radius $r$ of distance-aware filter $\psi_r$ is $12$. All codes are implemented with PyTorch on a single NVIDIA RTX 3090 GPU with 24GB memory.

 \vspace{-0.3cm}
\subsection{Experimental Results}
 \vspace{-0.2cm}
\label{sec:state-of-the-art}
We compare our method with multiple state-of-the-art methods: fAnoGAN \cite{schlegl2019f}, AnoGAN \cite{schlegl2017unsupervised}, MemAE \cite{gong2019memorizing}, WDMT-Net \cite{zhang2022multi}, Auto-Encoder \cite{baur2018deep}. Besides, we also transfer the representation-based state-of-the-art method in natural images PatchCore \cite{roth2022towards}  as the experimental baseline. As shown in Tab.\ref{tab:result_pixel} and Tab.\ref{tab:result_image}, ReSAD outperforms all baseline in pixel-level anomaly detection for both IDRiD and ADAM datasets and also reach state-of-the-art performance for the image-level result on IDRiD dataset.

Fig.\ref{fig:vis} shows the visualization of the abnormal score map. As we can observe, the network has a high degree of activation for the abnormal regions. Besides, as we can see from the figure,  ReSC can reduce the false positives in normal retinal structure regions and has better visualization for anomaly detection in fundus images compared with no ReSC module.

\begin{center}
\begin{table}[t]
\centering
\begin{tabular}{c|c|c|c|c}
\hline
Method & \multicolumn{2}{c|}{IDRiD \cite{porwal2020idrid}}  & \multicolumn{2}{c}{ADAM \cite{fang2022adam} } \\ \cline{2-5}
& AUC & ACC & AUC & ACC \\\Xhline{1.5 pt} \Xhline{0.5 pt}
Auto-Encoder \cite{baur2018deep} & 0.687 & 0.638 & 0.662 & 0.633\\
fAnoGAN \cite{schlegl2019f} & 0.729 & 0.715 & 0.716 & 0.693 \\ 
AnoGAN \cite{schlegl2017unsupervised} & 0.635 & 0.659 & 0.677 & 0.661 \\
MemAE \cite{gong2019memorizing} & 0.677 & 0.630 & 0.670 & 0.668 \\
WDMT-Net \cite{zhang2022multi} & 0.687 & 0.638 & 0.668 & 0.653 \\
PatchCore \cite{roth2022towards} & 0.845 & 0.758 & 0.733 & 0.678 \\
Ours & \textbf{0.907} & \textbf{0.819} & \textbf{0.820} & \textbf{0.732} \\
\hline
\end{tabular}
\vspace{-0.5em}
\caption{Pixel-level result for IDRiD and ADAM.}
\vspace{-0.5em}
\label{tab:result_pixel}
\end{table}
\end{center}
\vspace{-2.5em}
\subsection{Ablation}
\label{sec:ablation}
In the ablation experiment, we investigate the importance of the ReSC module (\S \ref{equ:feature_map}) by evaluating the performance without ReSC module, without region-aware feature
module, without spatial-aware feature module and with ReSC module on the IDRiD dataset. As shown in Tab.\ref{tab:ablation}, Compared with no ReSC module, the image-level AUC increases by 17.2\%, ACC increases by 14.6\%, pixel-level AUC increases by 4.3\%, and ACC increases by 4.2\%, proving that the ReSC module significantly improves the performance of anomaly detection in fundus images. 

Besides, to prove the data efficiency of ReSAD, we reduce normal images in the IDRiD training set. As shown in Tab.\ref{tab:ablation}, the reduction causes little performance degradation. Even utilizing only 10\% of the training data, ReSAD reaches 91.3\% AUC and 85.5\% ACC for image-level anomaly detection, 89.8\% AUC and 80.8\% ACC for pixel-level anomaly detection, which can still outperform most baseline results.

\begin{center}
\begin{table}[t]
\centering
\begin{tabular}{c|c|c}
\hline
Method & AUC& ACC \\\Xhline{1.5 pt} \Xhline{0.5 pt}

Auto-Encoder \cite{baur2018deep}  & 0.818 & 0.782 \\
WDMT-Net \cite{zhang2022multi} & 0.680 &  0.653 \\
MemAE \cite{gong2019memorizing}  & 0.722 & 0.689 \\
PatchCore \cite{roth2022towards} & 0.777 & 0.738 \\ 
fAnoGAN \cite{schlegl2019f} & 0.780 & 0.790 \\ 
Ours & \textbf{0.941} & \textbf{0.902} \\
\hline
\end{tabular}
\caption{Image-level result for IDRiD.}
\label{tab:result_image}
\end{table}
\end{center}

\begin{center}
\begin{table}[t]
\centering
\begin{tabular}{c|c|c|c|c}
\hline
Ablation & \multicolumn{4}{c}{IDRiD \cite{porwal2020idrid}} \\ \cline{2-5}
&I-AUC& I-ACC & P-AUC & P-ACC \\\Xhline{1.5 pt} \Xhline{0.5 pt}

10\%  & 0.913 & 0.855 & 0.898 & 0.808\\
40\% & 0.916 &  0.840 & 0.899 & 0.809\\
70\%  & 0.925 & 0.870 & 0.901 & 0.812 \\
\Xhline{1 pt}
wo ReSC & 0.769 & 0.756 &  0.864 & 0.777\\ 
wo Region Module & 0.812 & 0.785 & 0.878 & 0.792\\
wo Spatial Module & 0.915 & 0.848 & 0.897 & 0.807\\
with ReSC & 0.941 & 0.902 & 0.907 & 0.819 \\
\hline
\end{tabular}
\caption{Ablation study for training data number and ReSC. I-AUC, I-ACC denote Image-AUC and Image-ACC,  P-AUC, P-ACC denote Pixel-AUC and Pixel-ACC.}
\label{tab:ablation}
\end{table}
\end{center}
 \vspace{-2.2cm}
\section{CONCLUSION}
\label{sec:conclusion}
In this paper, we proposed a novel anomaly detection method ReSAD for fundus images. ReSAD adapted a pre-trained ImageNet model to extract the features and designed a  Region-and-Spatial-aware feature Combination module (ReSC) to construct the memory bank. Finally, ReSAD used the memory bank to give image-level and pixel-level anomaly scores. By evaluating our method on ADAM and IDRiD,  our method significantly outperformed the state-of-the-art anomaly detection methods, which proved the effectiveness of ReSAD in fundus abnormality detection.  

% Below is an example of how to insert images. Delete the ``\vspace'' line,
% uncomment the preceding line ``\centerline...'' and replace ``imageX.ps''
% with a suitable PostScript file name.
% -------------------------------------------------------------------------

% \vfill
% \pagebreak
\section{Compliance with Ethical Standards}
This research study was conducted retrospectively using human subject data available in the open access \textbf{IDRiD} and \textbf{ADAM} datasets. Ethical approval was not required as confirmed by the license attached with the open access data
\bibliographystyle{IEEEbib}
\bibliography{strings,refs}

\begin{thebibliography}{10}

\bibitem{shen2017deep}
Dinggang Shen, Guorong Wu, and Heung-Il Suk,
\newblock ``Deep learning in medical image analysis,''
\newblock {\em Annual review of biomedical engineering}, vol. 19, pp. 221,
  2017.

\bibitem{cai2022dual}
Yu~Cai, Hao Chen, Xin Yang, Yu~Zhou, and Kwang-Ting Cheng,
\newblock ``Dual-distribution discrepancy for anomaly detection in chest
  x-rays,''
\newblock {\em arXiv preprint arXiv:2206.03935}, 2022.

\bibitem{zhou2021proxy}
Kang Zhou, Jing Li, Weixin Luo, Zhengxin Li, Jianlong Yang, Huazhu Fu, Jun
  Cheng, Jiang Liu, and Shenghua Gao,
\newblock ``Proxy-bridged image reconstruction network for anomaly detection in
  medical images,''
\newblock {\em IEEE Transactions on Medical Imaging}, vol. 41, no. 3, pp.
  582--594, 2021.

\bibitem{salehi2021multiresolution}
Mohammadreza Salehi, Niousha Sadjadi, Soroosh Baselizadeh, Mohammad~H Rohban,
  and Hamid~R Rabiee,
\newblock ``Multiresolution knowledge distillation for anomaly detection,''
\newblock in {\em Proceedings of the IEEE/CVF conference on computer vision and
  pattern recognition}, 2021, pp. 14902--14912.

\bibitem{roth2022towards}
Karsten Roth, Latha Pemula, Joaquin Zepeda, Bernhard Sch{\"o}lkopf, Thomas
  Brox, and Peter Gehler,
\newblock ``Towards total recall in industrial anomaly detection,''
\newblock in {\em Proceedings of the IEEE/CVF Conference on Computer Vision and
  Pattern Recognition}, 2022, pp. 14318--14328.

\bibitem{schlegl2019f}
Thomas Schlegl, Philipp Seeb{\"o}ck, Sebastian~M Waldstein, Georg Langs, and
  Ursula Schmidt-Erfurth,
\newblock ``f-anogan: Fast unsupervised anomaly detection with generative
  adversarial networks,''
\newblock {\em Medical image analysis}, vol. 54, pp. 30--44, 2019.

\bibitem{zhang2022multi}
Wentian Zhang, Xu~Sun, Yuexiang Li, Haozhe Liu, Nanjun He, Feng Liu, and Yefeng
  Zheng,
\newblock ``A multi-task network with weight decay skip connection training for
  anomaly detection in retinal fundus images,''
\newblock in {\em International Conference on Medical Image Computing and
  Computer-Assisted Intervention}. Springer, 2022, pp. 656--666.

\bibitem{ouardini2019towards}
Khalil Ouardini, Huijuan Yang, Balagopal Unnikrishnan, Manon Romain, Camille
  Garcin, Houssam Zenati, J~Peter Campbell, Michael~F Chiang, Jayashree
  Kalpathy-Cramer, Vijay Chandrasekhar, et~al.,
\newblock ``Towards practical unsupervised anomaly detection on retinal
  images,''
\newblock in {\em Domain Adaptation and Representation Transfer and Medical
  Image Learning with Less Labels and Imperfect Data}, pp. 225--234. Springer,
  2019.

\bibitem{das2022anomaly}
Sourya~Dipta Das, Saikat Dutta, Nisarg~A Shah, Dwarikanath Mahapatra, and
  Zongyuan Ge,
\newblock ``Anomaly detection in retinal images using multi-scale deep feature
  sparse coding,''
\newblock in {\em 2022 IEEE 19th International Symposium on Biomedical Imaging
  (ISBI)}. IEEE, 2022, pp. 1--5.

\bibitem{sener2017active}
Ozan Sener and Silvio Savarese,
\newblock ``Active learning for convolutional neural networks: A core-set
  approach,''
\newblock {\em arXiv preprint arXiv:1708.00489}, 2017.

\bibitem{porwal2020idrid}
Prasanna Porwal, Samiksha Pachade, Manesh Kokare, Girish Deshmukh, Jaemin Son,
  Woong Bae, Lihong Liu, Jianzong Wang, Xinhui Liu, Liangxin Gao, et~al.,
\newblock ``Idrid: Diabetic retinopathy--segmentation and grading challenge,''
\newblock {\em Medical image analysis}, vol. 59, pp. 101561, 2020.

\bibitem{fang2022adam}
Huihui Fang, Fei Li, Huazhu Fu, Xu~Sun, Xingxing Cao, Fengbin Lin, Jaemin Son,
  Sunho Kim, Gwenole Quellec, Sarah Matta, et~al.,
\newblock ``Adam challenge: Detecting age-related macular degeneration from
  fundus images,''
\newblock {\em IEEE Transactions on Medical Imaging}, 2022.

\bibitem{mower2009prep}
Jeffrey~P Mower,
\newblock ``The prep suite: predictive rna editors for plant mitochondrial
  genes, chloroplast genes and user-defined alignments,''
\newblock {\em Nucleic acids research}, vol. 37, no. suppl\_2, pp. W253--W259,
  2009.

\bibitem{zagoruyko2016wide}
Sergey Zagoruyko and Nikos Komodakis,
\newblock ``Wide residual networks,''
\newblock {\em arXiv preprint arXiv:1605.07146}, 2016.

\bibitem{schlegl2017unsupervised}
Thomas Schlegl, Philipp Seeb{\"o}ck, Sebastian~M Waldstein, Ursula
  Schmidt-Erfurth, and Georg Langs,
\newblock ``Unsupervised anomaly detection with generative adversarial networks
  to guide marker discovery,''
\newblock in {\em International conference on information processing in medical
  imaging}. Springer, 2017, pp. 146--157.

\bibitem{gong2019memorizing}
Dong Gong, Lingqiao Liu, Vuong Le, Budhaditya Saha, Moussa~Reda Mansour, Svetha
  Venkatesh, and Anton van~den Hengel,
\newblock ``Memorizing normality to detect anomaly: Memory-augmented deep
  autoencoder for unsupervised anomaly detection,''
\newblock in {\em Proceedings of the IEEE/CVF International Conference on
  Computer Vision}, 2019, pp. 1705--1714.

\bibitem{baur2018deep}
Christoph Baur, Benedikt Wiestler, Shadi Albarqouni, and Nassir Navab,
\newblock ``Deep autoencoding models for unsupervised anomaly segmentation in
  brain mr images,''
\newblock in {\em International MICCAI brainlesion workshop}. Springer, 2018,
  pp. 161--169.

\end{thebibliography}

\end{document}